\pdfoutput=1
\documentclass[10pt,twocolumn,letterpaper]{article}

\usepackage{cvpr}
\usepackage{times}
\usepackage{epsfig}
\usepackage{graphicx}
\usepackage{amsmath}
\usepackage{amssymb}
\usepackage{cases}
\usepackage{multirow}
\usepackage{makecell}

\usepackage[lined,boxed,commentsnumbered]{algorithm2e}

\cvprfinalcopy % *** Uncomment this line for the final submission

\def\cvprPaperID{****} % *** Enter the CVPR Paper ID here

\begin{document}

%%%%%%%%% TITLE
\title{DHARI Report to EPIC-Kitchens 2020 Object Detection Challenge}

\author{Kaide Li, Bingyan Liao, Laifeng Hu, Yaonong Wang\\
%ZheJiang Dahua Technology CO.,LTD. \\
ZheJiang Dahua Technology CO.,LTD, Hangzhou, China\\
{\tt\small \{li\_kaide, liao\_bingyan, hu\_laifeng, wang\_yaonong\}@dahuatech.com}
% To save space, use either the email address or home page, not 
%\and
%Laifeng Hu\\
%\and
%Laifeng Hu\\
}

\maketitle
%\thispagestyle{empty}

%%%%%%%%% ABSTRACT
\begin{abstract}
%	In this report we describe the technical details of our	submission to the EPIC-Kitchens Object Detection Challenge.	
%	We first introduce duck filling and mix-up technique to expand the data and proved to significantly improve the robust of our methods.
%%	The duck filling and mix-up are utilized to further expand the data and proved to be helpful to this mission. 
%	Then we propose GRE-FPN and Hard IoU-imbalance Sampler methods to further obtain a better performance. And to bridge this gap of category imbalance, Class Balance Sample is utilized and greatly improve the test results. 
%	Finally, through Stochastic Weight Averaging and multi scale test, we get a better and more robust test results on benchmarks. Experimental results demonstrate our approach can significantly improve the mean Average Precision (mAP) of object detection on both the seen and unseen test sets of EPIC-Kitchens.

In this report, we describe the technical details of our submission to the EPIC-Kitchens Object Detection Challenge. Duck filling and mix-up techniques are firstly introduced to augment the data and significantly improve the robustness of the proposed method. Then we propose GRE-FPN and Hard IoU-imbalance Sampler methods to extract more representative global object features. To bridge the gap of category imbalance, Class Balance Sampling is utilized and greatly improves the test results. 
Besides, some training and testing strategies are also exploited, such as Stochastic Weight Averaging and multi-scale testing. 
%And in training and testing phase we also exploit some little tips, such as Stochastic Weight Averaging and multi-scale testing.
%Finally, through Stochastic Weight Averaging and multi-scale test, we get better and more robust test results on benchmarks. 
Experimental results demonstrate that our approach can significantly improve the mean Average Precision (mAP) of object detection on both the seen and unseen test sets of EPIC-Kitchens.
\end{abstract}

%%%%%%%%% BODY TEXT
\section{Introduction}

%In this report, we will describe the technical details of our submission to the EPIC-Kitchens 2020 objection detection challenge~\cite{EPIC-kitchen}. We evaluate several different networks and training strategies. Additionally, we also try to balance the influence of the long-tail class distribution and improve the diversity of few-shot classes to make the training set more balanced among all categories to obtain a desired result. Experimental results demonstrate our approach can significantly improve the object detection performance.

EPIC-Kitchens dataset was introduced as a large scale first-person action recognition dataset. In this Object detection task, the annotations only capture the 'active' objects pre-, during- and post- interaction~\cite{EPIC-kitchen}~\cite{epicdataset}.
It is challenging for object detection due to the influence of the sparse annotations and long-tail class distribution in this dataset.
%Detection in this dataset is challenged by the influence of the sparse annotations and long-tail class distribution.
To address these challenges, we focus on the sampling methods in detection process and present the GRE-FPN and Hard IoU-imbalance Sampler methods to improve the robustness of location. Additionally, duck filling methods are applied to balance the influence of the long-tail class distribution and improve the diversity of few-shot classes.
Experimental results demonstrate our approach can significantly improve the object detection performance and
achieve a competitive result on the test set.
%and achieves mAP 41.72\% for Seen Kitchens (S1) and 39.93\% for Unseen Kitchens (S2) with $IoU\textgreater0.5$.
The implementations details of the above are described in section 2 and section 3.

%-------------------------------------------------------------------------
\section{Proposed Approach}

\subsection{Data Preprocess}
\label{datapro}

\begin{figure}[t]
	\begin{center}
		\includegraphics[width=0.95\linewidth]{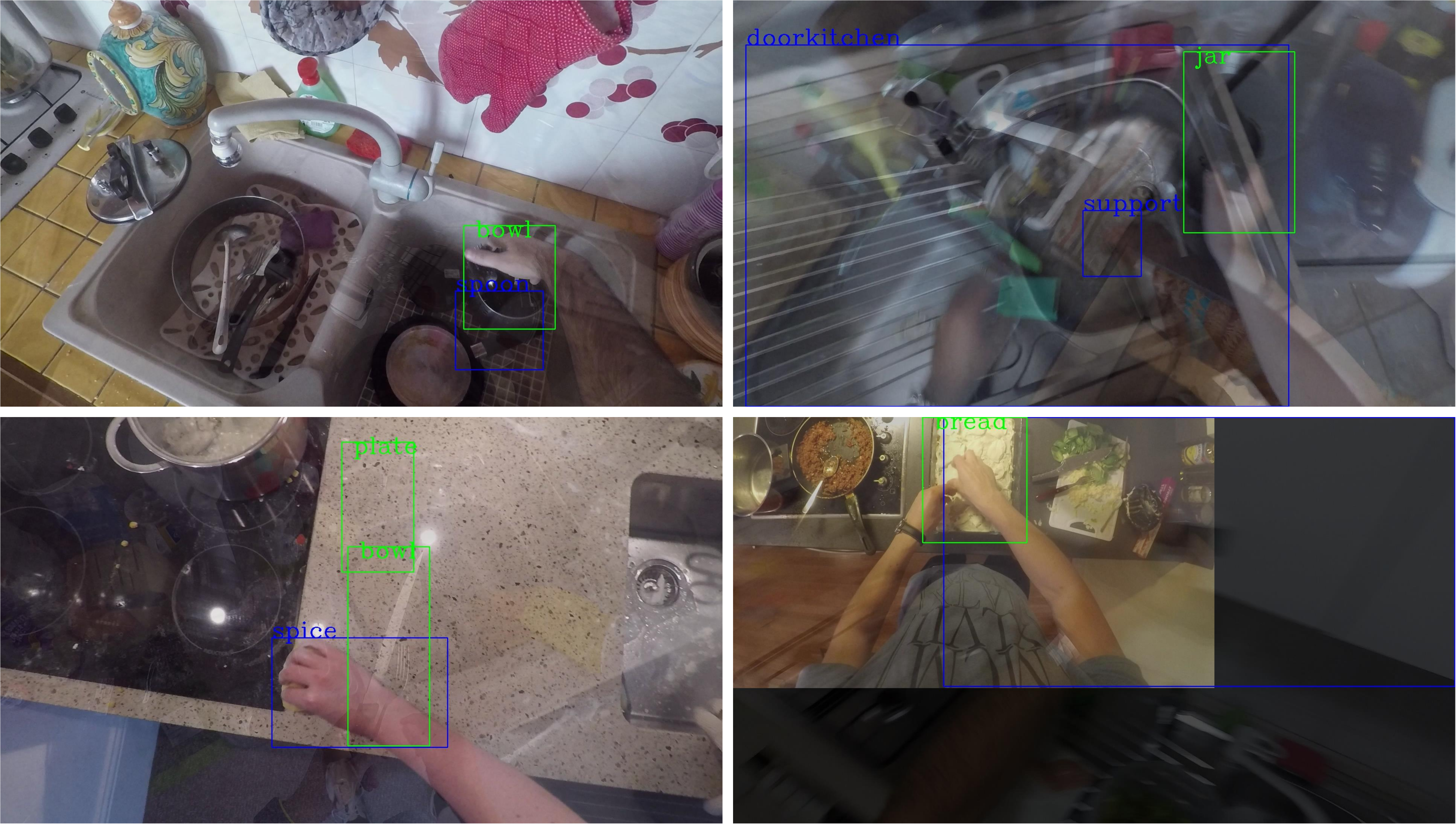}
	\end{center}
	\caption{The mix-up images for few-shot classes.}
	\label{mix_up}
\end{figure}

\begin{figure}[t]
	\begin{center}
		\includegraphics[width=0.95\linewidth]{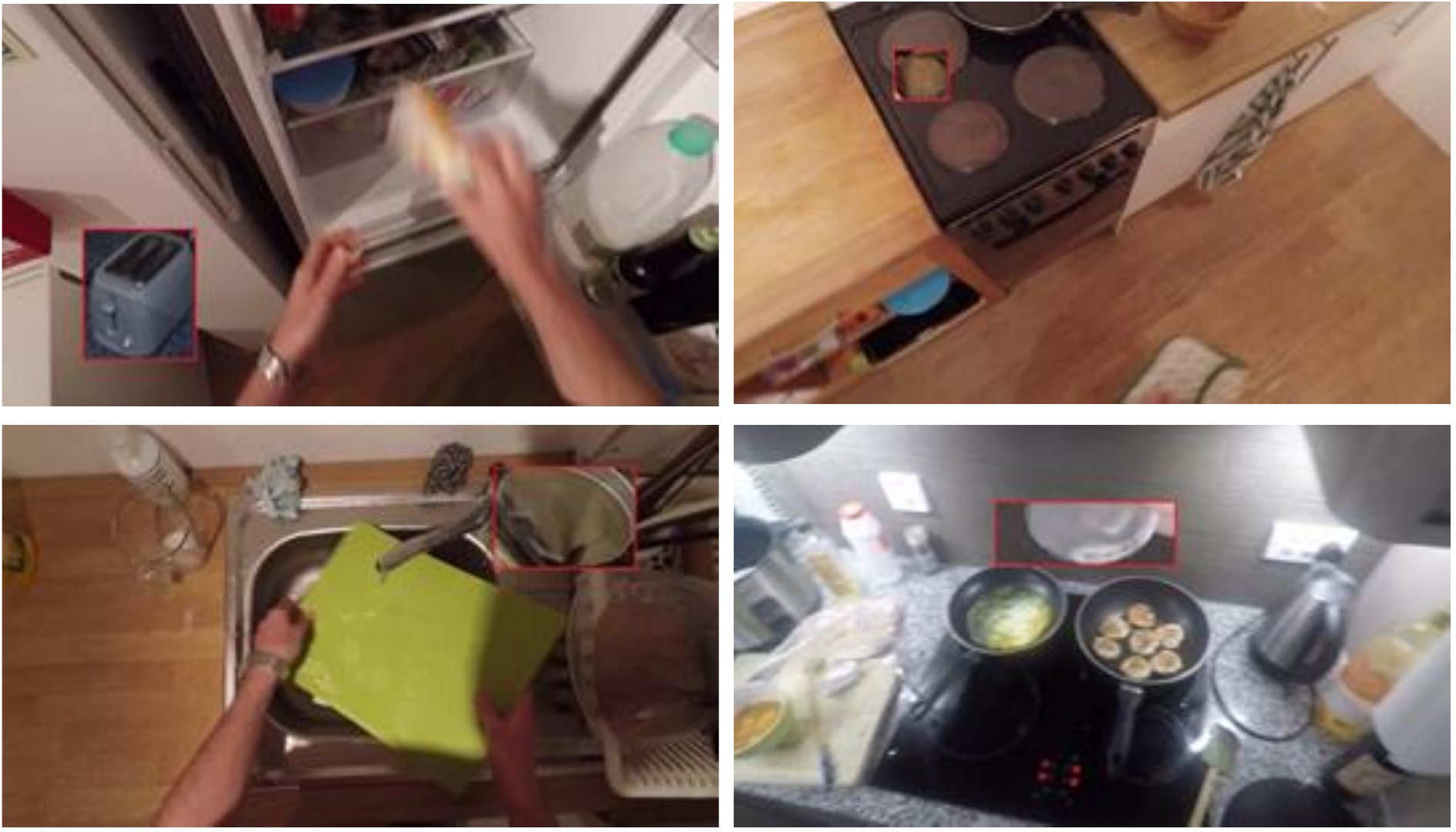}
	\end{center}
	\caption{The duck filling images for few-shot classes.}
	\label{duck_fill}
\end{figure}
The EPIC-Kitchens training dataset for object detection contains 290 valid annotation categories and 326,064 bounding boxes. But the training dataset is extremely imbalanced, and it has the long-tail class distribution. Especially, there are only 8,136 bounding boxes for few-shot classes (118 classes) with less than 200 bounding boxes in total. To address the imbalance between few-shot classes and many-shot classes and improve the diversity of few-shot classes, we adopt some data augmentation preprocess, such as mix-up~\cite{mixup} and few-shot classes duck filling.
For mix-up, it fuses two visually coherent images together into one output image by using Beta random distribution to improve the diversity of training set. And the mixtures are present in the Fig~\ref{mix_up}.
For few-shot classes duck filling, we first extract the few-shot bounding boxes from training set, and then fill the few-shot objects into the non-annotated images in training set. In filling processing, some skills such as random weighted average method, random rescale for bounding boxes, are utilized to realize the mixing between few-shot objects and non-annotated images. The result images for few-shot classes duck filling are shown in Fig~\ref{duck_fill}. 
Besides the two specific data argument methods, some regular augment methods are applied, such as random scale, random flipping, channel shuffling and random brightness contrast and so on.

\subsection{Proposed Methods}
\label{methods}
We exploit both one-stage (such as FCOS~\cite{fcos:}, ATSS~\cite{ATSS}) and two-stage (Cascade-RCNN~\cite{cascade_rcnn}) as the basic detectors and evaluate their performance on validation dataset. And several classification networks are chosen as the backbone, such as HRNet-w36, HRNet-w48~\cite{hrnet}, ResNext101-64, ResNext101-32~\cite{resnext} and ResNet101~\cite{resnet}. 
Comparing the detection performance of the above methods, we select the Cascade-RCNN as base network framework. And The ResNet101 is chosen as the backbone with FPN and deformable convolution (DCN).
%Comparing the detection performance of the above frames, we choose Cascade-RCNN with backbone of ResNet101 as a base network framework.
Besides, two other improvement skills are introduced to obtain a better performance --- GRE-FPN and Hard IoU-imbalance Sampler. 

\begin{figure}[t]
	\begin{center}
		\includegraphics[width=0.95\linewidth]{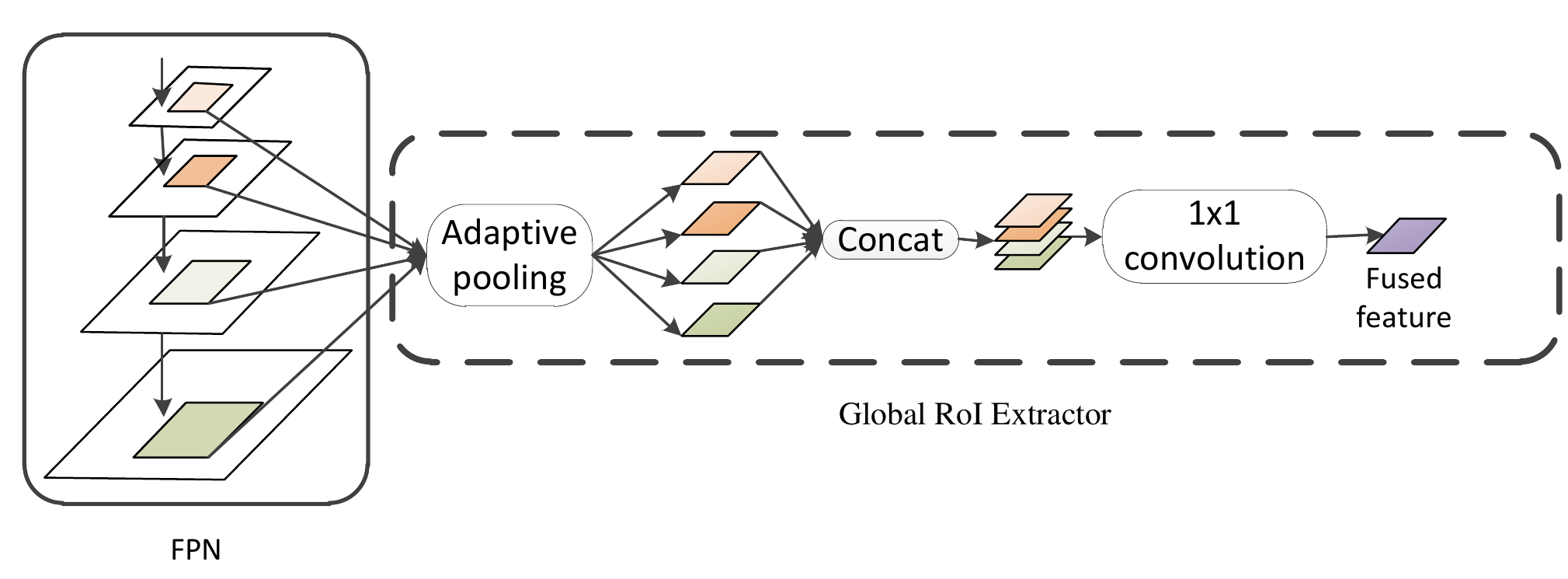}
	\end{center}
	\caption{The details of Global RoI Extractor}
	\label{GRE}
\end{figure}

\textbf{GRE-FPN:} At the regular FPN~\cite{fpn} stage, RoI features are usually extracted from one certain pyramid level according to the scales of RoIs. It ignores the importance of the adjacent scale features, which may contain more accurate location information. Therefore, we propose the Global RoI Extractor (GRE), which can extract RoI features from all pyramid levels and learn adaptive weights to balance the importance of features on different pyramid levels automatically. The detailed structure of GRE is illustrated in Fig~\ref{GRE}. We first pool RoI features from all pyramid levels and concatenate them together. Then, these pooled features are convoluted with 1x1 convolution to reduce the channel dimension and obtain the final RoI features to predict objects.

\begin{figure}[t]
	\begin{center}
		\includegraphics[width=0.95\linewidth]{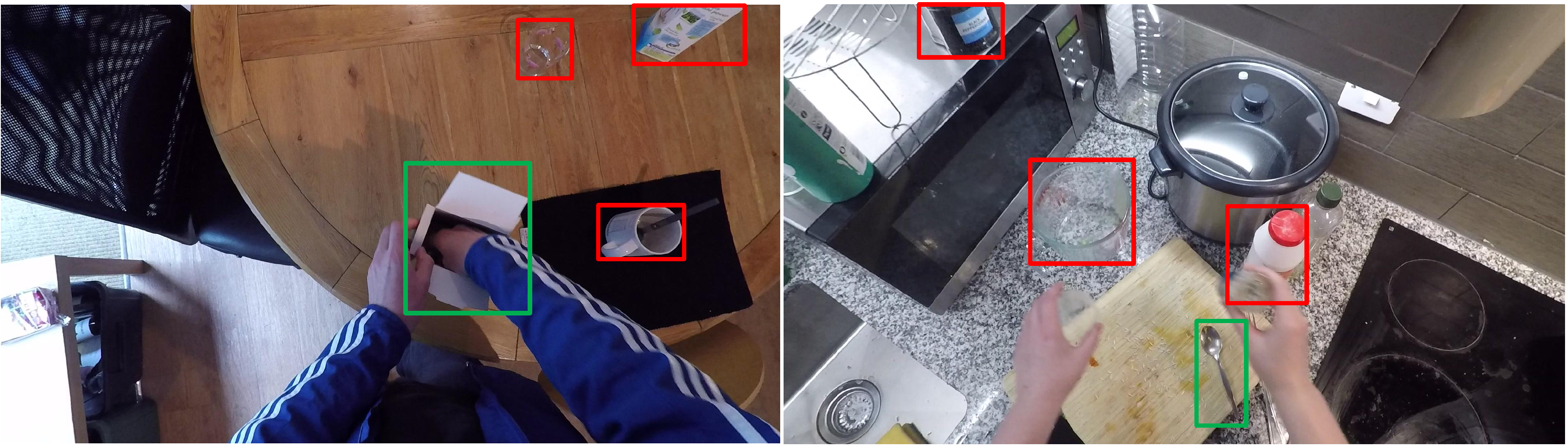}
	\end{center}
	\caption{The annotation visualize examples}
	\label{iou_ba}
\end{figure}

\textbf{Hard IoU-imbalance Sampler:} We visualize annotations of objects on the respective images and find the annotations are sparse and coarse annotated. Besides, amounts of noise information are introduced and many targets should be labeled are missing. We call these lost targets unmarked targets. As for anchor-based network, the qualities of proposed anchors are significant for achieving an outstanding performance. 
Considering these images shown in Fig~\ref{iou_ba}, we can observe that the marked target (the green box) may be surrounded by other unmarked targets (the red boxes). 

The regular sampling methods, such as Random Sampling, Online Hard Example Mining (OHEM), may generate false samples that take the unmarked target as negative samples and lead to the decreased robustness at the stage of RPN.
To reduce the possibility of taking the unmarked targets as negative samples, we constrain negative sample regions that IoU ($ IoU_{(S,T)} $) and center distance with marked target ($ DIS_{(S, T)} $) are subject to
%we choose negative samples whose IoU and center distance with marked target are subject to

\begin{equation}
\begin{cases}
IoU_{(S,T)} &<0.3 \\
DIS_{(S,T)} &<â€?T_w, T_h)‖_2
\end{cases}
\end{equation}

$ S $ is the proposal anchor, and $ T $ is the annotated target bounding box. $ T_w $, $ T_h $ are the width and height of annotated target respectively. In this methods, the negative sample region always surround the marked targets rather than the whole input image, which can significantly reduce the influence of false sample. 
Besides, inspired by~\cite{Pang2019Libra}, we also improve the sampling ratio of hard negative examples with IoU in range $ (0.05, 0.3) $ to further ensure the balance of sample processing. 

%\section{3333333}

\section{Experiments}

In this section, we conduct several experiments on EPIC-Kitchens object detection dataset. And the comparisons of detection performance are presented to verify the effectiveness of proposed method.

%In this section, we examine some training and testing strategies which affect the performance of the model with the same backbone networks. The compared results on validation datasets are shown in Table~\ref{train_comp}. Our proposed strategies can better solve the three major problems of the EPIC Datasets: the long-tail class distribution, the domain gap between the seen and unseen test sets, and missing annotations of the bounding boxes of objects.

\begin{table*}[t]
	\begin{center}
		\begin{tabular}{cccccccc}
			\hline
			Cascade-rcnn & EPIC-Pretrain & CB & Multi-Test & SWA & IoU\textgreater0.05(\%) & IoU\textgreater0.5(\%) & IoU\textgreater0.75(\%)  \\
			\hline
			\checkmark &  &  &  &  & 71.11 & 43.22 & 17.12\\               
			\checkmark & \checkmark  &  &  &  & 72.01 & 43.67 & 18.01\\ 
			\checkmark &  & \checkmark  &  &  & 72.77 & 44.67 & 18.55 \\
			\checkmark &  &  & \checkmark  &  & 72.45 & 44.51 & 18.32\\         
			\checkmark &  &  &  & \checkmark  & 71.97 & 43.78 & 17.89\\     
			\hline   
			\checkmark & \checkmark  & \checkmark  &  &  & 73.54 & 45.10 & 18.97\\
			\checkmark & \checkmark  & \checkmark  & \checkmark  &  & 73.87 & 45.57 & 19.22\\
			\checkmark & \checkmark  & \checkmark  & \checkmark  & \checkmark  &  \textbf{74.64} &  \textbf{46.22} &  \textbf{20.13}\\
			%        \checkmark & \checkmark  & \checkmark  & \checkmark  & \checkmark  &  &  & \\
			%        \checkmark & \checkmark  & \checkmark  & \checkmark  & \checkmark  &  &  & \\
			
			\hline
		\end{tabular}
	\end{center}
	\caption{Comparison results on EPIC validation dataset. Performance measures contain mAP with ratio 0.05, 0.5, 0.75.}
	\label{train_comp}
\end{table*}

\begin{table*}[t]
	\small 
	\begin{center}
		\begin{tabular}{c|c|c|c|c|c|c|c|c|c|c}
			\hline
			& \multirow{2}*{method} &  \multicolumn{3}{c|}{Few shot classes(\%)} &  \multicolumn{3}{c|}{Many shoy classes(\%)} & \multicolumn{3}{c}{All classes(\%)} \\    
			\cline{3-11}
			&  & \makecell{IoU\textgreater\\0.05} & \makecell{IoU\textgreater\\0.5} & \makecell{IoU\textgreater\\0.75} & \makecell{IoU\textgreater\\0.05} & \makecell{IoU\textgreater\\0.5} & \makecell{IoU\textgreater\\0.75} &\makecell{IoU\textgreater\\0.05} & \makecell{IoU\textgreater\\0.5} & \makecell{IoU\textgreater\\0.75}\\		
			\hline	 	
			\multirow{6}*{seen} & base & 47.52 & 26.01 &7.56 & 36.10& 39.59 &10.79 & 60.17& 37.03 &10.18 \\ 
			\cline{2-11}		
			& Mixup + duck & 49.86 &31.43  & 12.39 & 64.01 & 40.26 & 13.24 & 61.35 & 38.60 & 13.08\\ 
			\cline{2-11}		
			& Mixup + duck + Train tricks & 47.07 & 27.01 & 11.79 & 65.68 & 41.72 & 13.92 & 62.18 & 38.95 &13.52\\ 
			\cline{2-11}		
			& Mixup + duck + Hard IoU & 52.48 & \textbf{33.30} & 13.29 & 67.27 & 41.16 & 12.91 & 65.23 & 39.68 &12.98\\ 
			\cline{2-11}		
			& Mixup + duck + GRE-FPN & 47.14 & 32.02 & 13.35 & 66.23 & 42.47 & 14.52 & 53.58 & 39.83 &14.30\\ 
			\cline{2-11}
			& Ensemble &\textbf{ 54.98} & 32.40 & \textbf{ 14.55} &  \textbf{68.74} &  \textbf{43.88} &  \textbf{15.38} &  \textbf{66.15} &  \textbf{41.72} &  \textbf{15.23}\\ 																				           
			\hline	
			\multirow{6}*{unseen}& base & 20.07 & 13.42 & 1.93 & 54.83 & 31.92 &8.11 & 51.29 & 30.04 &7.48\\    
			\cline{2-11}		
			& Mixup + duck &  \textbf{42.95} & 18.26 & 4.86 & 65.42 & 38.24 & 11.68 & 63.19 & 36.20 & 10.99\\ 
			\cline{2-11}		
			& Mixup + duck + Train tricks & 39.44 &  \textbf{24.20} &  \textbf{8.61} & 66.44 & 38.85 & 12.79 & 63.69 & 37.36 & 12.36\\ 
			\cline{2-11}		
			& Mixup + duck + Hard IoU & 40.99 & 22.66 & 6.82 & 66.62 & 39.13 & 12.42 & 64.01 & 37.45 & 11.85\\ 
			\cline{2-11}		
			& Mixup + duck + GRE-FPN & 34.02 & 19.42 & 7.52 & 66.81 & 39.71 & 13.38 & 63.47 & 37.64 & 12.78\\ 
			\cline{2-11}
			& Ensemble & 35.75 & 22.31 & 7.33 &  \textbf{67.92} &  \textbf{41.92} &  \textbf{14.29} &  \textbf{64.64} &  \textbf{39.93} &  \textbf{13.58} \\ 																				           
			\hline	
		\end{tabular}
	\end{center}
	\caption{Comparison results on EPIC testing dataset on mAP. Ensemble results based on all designed variability. Challenge website details: https://competitions.codalab.org/competitions/20111. Note that our best model was submitted under the anonymous nickname DH-ARI}
	\label{final_res}
\end{table*}

\subsection{Experimental Settings}

All the experiments are conducted by using the MMdetection toolbox which is developed with PyTorch by Multimedia Laboratory, CUHK. And we run our experiments on 8 NVIDIA P40 GPUs.
The mini-batch Stochastic Gradient Descent (SGD) optimizer with momentum of 0.9 and weight decay of $ 1\times e^{-4} $  is utilized to solve the experiments. The input images are randomly sized to ($ 1280\times 720$) and ($ 1394\times 764$). The batch size is 32 and the maximum epoch for training is set to 12. The initial learning is fixed to 0.02. Then, it decays to $ 2\times e^{-3} $ at epoch 8 and $ 2\times e^{-4} $ at epoch 11. Meanwhile, we use 0.0067 to warm up the training until 500 iterations, then go back to 0.02 and continue training.

%In practice, we select the Cascade-RCNN as final network framework. And The ResNet101 is chosen as the backbone with FPN and deformable convolution (DCN).
According to the number of bounding boxes for each category, the whole training set is divided into two parts : many shot (S1), few shot (S2). In the entire training phase, there are three stages: T1, T2, T3. In training phase T1, we choose datasets S1 to train and get a trained model (M1); 
In T2, we choose datasets S2 to fine-tune model M1 and obtain fine-tuning model (M2); 
And in the final phase T3, we use both datasets S1 and S2 to fine-tuning model M2 and obtain the final model.

\subsection{Training Skills}
\label{training skill}
In training and validation stage, we use several training skills to optimize the training procedure, such as Model Pre-training, Class Balance Sampling and Stochastic Weight Averaging. The test results on validation dataset is show in Table~\ref{train_comp}. The validation dataset is extracted from train dataset, and is divided into seen and unseen set.

\label{pretrained}
\textbf{Model Pre-training.} We extract images based on bounding boxes and categories of objects from EPIC-Kitchens Object Detection datasets, and train a Classification model as our pre-trained model to replace default ImageNet pre-trained model. 

\textbf{Class Balance Sampling.} During the training phase, image lists are randomly shuffled before the start of every epoch. Considering the long-tail class distribution and imbalanced categories, we randomly sample images in training list based on category probability as formula~(\ref{CB}). 
\begin{equation}
\begin{cases}
W_i&= 1/S_{i_c} \\ %\sum_{c=0}^m{(\mathbb{R}_{(c, i_c)}*S_c)} \\
S_{i_c}&= \sum_{j=0}^m{\mathbb{R}_{(c,c_{i_j})}}
\label{CB}
\end{cases}
\end{equation}
Where $ W_i $ and $ S_{i_c} $ is the sampling possibility and the total numbers of category id $ c $ in the $ i^{th} $ image in training list respectively. 
%$ c $ is the minimum class among all categories in image $ i $. 
$ c $ is the class ID that is annotated in image $ i $ and has the minimum proportion in the whole training dataset.
$ m $ represents the total marked categories in the image $ i $.
$ c_{i_j} $ is the category of the $ j^{th} $ object in image $ i $. And $ \mathbb{R}_{(c, c_{i_j})} = 1 $ only if $ c = c_{i_j} $, otherwise $ \mathbb{R}_{(c, c_{i_j})} = 0 $.
This sampling methods can increase the sampling possibility of the few shot class and effectively solve the imbalanced categories problem.

\textbf{Stochastic Weight Averaging (SWA).} SWA~\cite{Izmailov2018Averaging} is based on averaging the samples proposed by SGD with a learning rate schedule that allows exploration of the region of weight space corresponding to high-performing networks. Using SWA, we achieve notable improvement over conventional SGD training on our base model and this method has almost no computational overhead.

\subsection{Testing}

In testing phase, we adopted the fusing results of multi-scale~\cite{He2014Spatial}, flipping and Gaussian fuzzy transformation as the test output, which can further improve the detection performance.
%Besides,  are also applied to the testing processing. 
%By fusing the test results of multi-scale, flipping and Gaussian fuzzy transformation, we gain a distinct improvement on testing set. 
%By fusing the above strategies, the detection results are obtained and they are presented in Table~\ref{final_res}.
The detection results with different methods are presented in Table~\ref{final_res}.
%Compared with the single-scale test, the multi-scale test has a great improvement in effect~\cite{He2014Spatial}.Therefore, multi-scale testing is used as an effective technique to improve performance in competitions such as MS COCO. In our final submission result, we fuse the results of three scales, flip and Gaussian fuzzy transformation and gain a clear improvement in the final indicator. The final test results can be found in Table~\ref{final_res}. 
Specifically, by using data augmentation proposed in Sec~\ref{datapro}, proposed methods in Sec~\ref{methods} and training tricks in Sec~\ref{training skill}, 
the best performance obtained for seen and unseen results in a mAP of 39.93\% with $ IoU\textgreater 0.5 $. 
Using the duck filling technique, the recognition accuracy improved by 6.16\% (30.04\% vs 36.20\%) on unseen set. 
By combining the data argument with training tricks, Hard IoU and GRE-FPN models separately, improvements of 1.16\%, 1.25\%, 1.44\% are obtained respectively.
With an ensemble of all the methods, the mAP of $ IoU\textgreater 0.5 $ is further improved by 3.73\% (36.20\% vs 39.93\%).

\section{Conclusion}

The proposed method for the EPIC-Kitchens object detection task is demonstrated in detail in this paper. 
Our main concerns are to moderate the long-tail class distribution of training set and extract more effective features.

The duck filling, mix-up and class balance sampling are introduced to expand the training set and moderate the long-tail distribution. And The Hard IoU-imbalance Sampler and the reconstructed GRE-FPN are also utilized to help extracting more representative object features. 
%Class Balance Sampling are applied to bridge the gap of  category imbalance.
Experimental results demonstrate that our methods are effective and useful. By assembling these main methods, our detection framework can obtain a competitive performance on both the seen and unseen test data.

%We tried several detector frameworks and neural network backbones. The Cascade-RCNN with neural network backbone of ResNet101 is selected for better experimental results. Data augmentation of duck filling and mix-up is proved to be effective. Experimental results demonstrate that our methods GRE-FPN and Hard IoU-imbalance Sampler are effective and useful. Class Balance Sample is utilized to overcome the problem of the long-tail class distribution. Finally, we use Stochastic Weight Averaging and multi-scale test to obtain a final robust submission, and our model achieves quite competitive results on both the final seen and unseen test data leaderboards.

%{\small
%\bibliographystyle{ieee_fullname}
%\bibliography{egbib}
%}

\end{document}